%% file: mora_arxiv.tex
\documentclass[11pt,letterpaper]{article}
\usepackage[margin=1in]{geometry}
\usepackage[T1]{fontenc}
\usepackage{times}
\usepackage[authoryear,round]{natbib}
\setcitestyle{citesep={;},aysep={,},yysep={;}}
\usepackage{caption}
\usepackage[section]{placeins}
\input{math_commands.tex}

\usepackage{url}
\usepackage{booktabs}
\usepackage{multirow}
\usepackage{amsmath,amssymb}
\usepackage{graphicx}
\usepackage{xcolor}
\usepackage{microtype}
\usepackage{wrapfig}
\usepackage[hidelinks]{hyperref}
\hypersetup{
  pdftitle={MoRA: MoE Pruning via Router Bias Learning and Expert Approximation},
  pdfauthor={Yushuai Sun, Zikun Zhou, Lin Gao, Jun Yu, Wenjie Pei},
  pdfsubject={Preprint},
  pdfkeywords={mixture of experts, expert pruning, router bias, model compression}
}

\title{MoRA: MoE Pruning via Router Bias Learning and Expert Approximation}
\author{%
  \normalsize Yushuai Sun \quad
  Zikun Zhou\thanks{Corresponding authors}
  \quad Lin Gao \quad Jun Yu \quad Wenjie Pei\footnotemark[1]\\[0.4em]
  \normalsize Harbin Institute of Technology, Shenzhen\\[0.4em]
  \small \href{mailto:yushuai112233@gmail.com}{\nolinkurl{yushuai112233@gmail.com}}\quad
  \href{mailto:zhouzikunhit@gmail.com}{\nolinkurl{zhouzikunhit@gmail.com}}\\[0.2em]
  \small \href{mailto:gaol@hit.edu.cn}{\nolinkurl{gaol@hit.edu.cn}}\quad
  \href{mailto:yujun@hit.edu.cn}{\nolinkurl{yujun@hit.edu.cn}}\quad
  \href{mailto:wenjiecoder@outlook.com}{\nolinkurl{wenjiecoder@outlook.com}}
}
\date{}

\begin{document}

\maketitle

\begin{abstract}
Mixture-of-Experts (MoE) models enable parameter scaling with limited per-token computation by activating only a small subset of experts for each token, but deploying them still requires loading the complete expert pool into memory. Structured expert pruning can effectively reduce the memory usage by removing experts. However, existing pruning methods either use expert ranking criteria that are not well aligned with model performance or rely on effective expert subset searching that is computationally expensive. Moreover, these methods typically overlook the routing-behavior redundancy among the retained experts. In this paper, we propose MoE Pruning via Router Bias Learning and Expert Approximation (MoRA), a framework for structured MoE expert pruning. We introduce a learnable router bias for each expert and optimize these biases by minimizing the language-modeling loss and a routing-diversity regularizer. The learned router biases sharpen the routing probability distributions to identify experts critical to model performance while encouraging the selection of experts with diverse routing preferences. In addition, we introduce an expert approximation mechanism as a post-pruning enhancement. It leverages the remaining experts to approximate the outputs of pruned experts by affine transformation, further improving the performance of the pruned model. We evaluate MoRA on Qwen3-30B-A3B, DeepSeek-V2-Lite, and Moonlight-16B-A3B, removing 25\% and 50\% of the routed experts in each MoE layer. Extensive experiments on nine zero-shot benchmarks show that MoRA outperforms state-of-the-art pruning algorithms. Our code will be released.

\end{abstract}

\section{Introduction}
\label{sec:intro}
Mixture-of-Experts (MoE) models activate only a small subset of experts for each token, enabling model scaling with limited per-token computation \citep{shazeer2017outrageously,fedus2022switch}.
Although sparse activation reduces inference computation, deploying MoE models still requires loading the full expert pool, resulting in substantial memory usage.
Structured expert pruning offers a direct way to reduce this cost by removing less important experts while retaining the sparse routing structure \citep{lee2025stun,xie2024moe,mone2026}.
Hence, the core challenge is to determine which experts can be removed while preserving the predictive behavior of the model.

\begin{figure}[htbp]
\centering
\includegraphics[width=0.99\textwidth]{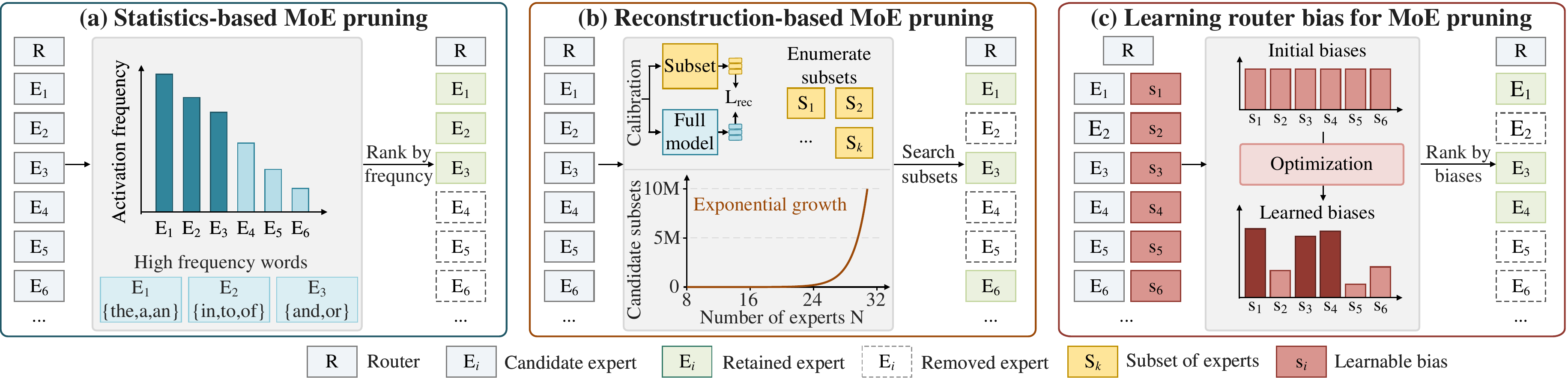}
\caption{Comparison of three ways to select retained MoE experts. (a) Statistic-based pruning assigns each expert a fixed score based on routing statistics and retains the highest-scoring experts. (b) Reconstruction-based pruning evaluates candidate expert subsets using layer-output reconstruction error and selects the subset with the lowest reconstruction error. (c) MoRA adds a router bias to each expert to sharpen the routing probability distribution. These biases are optimized to establish an expert ranking for subsequent selection.}
\label{fig:overview}
\end{figure}

Early expert pruning methods rank experts using aggregate routing statistics, such as activation frequency and average gate values \citep{koishekenov2023memory,xie2024moe}. However, these expert ranking criteria may not directly align with the model predictive performance. For example, frequency-based pruning may assign higher scores to experts that frequently process common function words with less semantic information \citep{dong2026domain}, as illustrated in Figure~\ref{fig:overview} (a).
Recent methods select an expert subset by minimizing layer-wise output reconstruction error on a calibration dataset under a prescribed retention budget \citep{lu2024not}, as illustrated in Figure~\ref{fig:overview} (b). However, these methods require evaluating candidate expert subsets through a set-level search, whose cost grows rapidly with the number of experts.
Moreover, the above-mentioned methods generally do not consider the diversity in the routing behavior of retained experts. Consequently, they may retain experts with highly similar routing patterns, leading to redundancy within the retained set.

To address these issues, we propose MoE pruning via Router bias learning and expert Approximation (MoRA), a framework for structured MoE expert pruning. As shown in Figure~\ref{fig:overview} (c), we first introduce a learnable router bias for each expert. These experts are originally pretrained under the load-balancing constraints. The biases can sharpen routing probability distributions over experts and serve as scores for selecting experts. To learn effective biases, we optimize them with the language-modeling (LM) loss and a routing-diversity regularizer.
The former encourages higher biases for experts critical to predictive performance, while the latter discourages experts with similar routing preferences from receiving high biases simultaneously. Herein, similar routing preference means that experts tend to receive similar routing weights for the same tokens.

We also introduce an expert approximation mechanism in our MoRA as a post-pruning enhancement. Specifically, we assign a retained expert as a proxy for each pruned expert and apply an affine transformation to approximate the output of the pruned expert with the retained experts, thereby mitigating the performance degradation caused by pruning. We evaluate MoRA on Qwen3-30B-A3B~\cite{qwen3report2025}, DeepSeek-V2-Lite~\cite{deepseekv22024}, and Moonlight-16B-A3B~\cite{liu2025muon} under two deployment budgets that remove 25\% and 50\% of the routed experts in each MoE layer. We evaluate zero-shot downstream performance on nine benchmarks. Further analysis reveals that the association between the learned router biases and predictive performance is stronger in models with more experts and in deeper MoE layers.

Our main contributions are as follows:
\begin{itemize}
    \item We propose MoRA, a structured MoE expert pruning framework that learns router biases using the LM loss and a routing-diversity regularizer. The learned biases serve as scores for selecting experts with high predictive importance and diverse routing preferences.

    \item We propose expert approximation as a post-pruning enhancement, which uses lightweight affine transformations to approximate the outputs of pruned experts with the retained experts. It effectively mitigates the performance degradation caused by pruning.

    \item We evaluate MoRA on Qwen3-30B-A3B, DeepSeek-V2-Lite, and Moonlight-16B-A3B at 25\% and 50\% routed-expert pruning. Results on nine zero-shot benchmarks show that MoRA outperforms other pruning baselines while better preserving the predictive performance.
\end{itemize}

\section{Related Work}
\label{sec:related}

\noindent\textbf{Sparse MoE models.}
Sparse MoE models increase model capacity while limiting per-token computation by activating only a subset of experts \citep{shazeer2017outrageously}. Large-scale systems such as GShard \citep{lepikhin2020gshard} and Switch Transformers \citep{fedus2022switch} established routing and training recipes for this regime. Despite sparse activation, storing all expert parameters remains memory-intensive. This motivates expert pruning to reduce memory requirements while preserving predictive performance.

\noindent\textbf{MoE Pruning.}
Weight-level pruning methods, such as SparseGPT, Wanda, and MoE-Pruner, can reduce MoE parameters by pruning individual weights within experts \citep{frantar2023sparsegpt,sun2024simple,xie2024moe}. However, these unstructured pruning methods directly reduce storage requirements and typically require specialized hardware support to achieve inference speedup.
Expert-level pruning instead removes entire experts, directly reducing storage requirements. Early methods rank experts using activation frequency and gating scores \citep{koishekenov2023memory}. NAEE \citep{lu2024not} selects expert subsets that minimize the reconstruction loss between the original and pruned layer outputs. However, enumerating the subsets leads to expensive computation as the number of experts and the pruning budget increase. STUN \citep{lee2025stun} avoids exhaustive subset evaluation through expert clustering and parameter-space approximations.
Complementary approaches compress MoE models through expert merging \citep{li2024merge,hcsmoe2025}, or replacement with lightweight modules \citep{mone2026,lightmoe2026}. However, modifying or replacing expert functions can introduce additional approximation errors.

\section{MoRA}
\label{sec:method}

In this section, we propose MoE pruning via Router bias learning and expert approximation (MoRA), a framework for structured MoE pruning. We introduce a learnable router bias for each expert to sharpen the routing probability distributions, and optimize these biases using the LM loss and a routing-diversity regularizer. The resulting biases serve as expert selection scores that balance predictive importance and routing diversity.

Then, MoRA performs expert approximation based on affine transformation as a post-pruning enhancement. Specifically, MoRA estimates the directed proxy fidelity between retained and removed experts. It fits a channel-wise affine transformation from each candidate retained-expert output to the corresponding removed-expert output and evaluates the resulting approximation quality. For each removed expert, MoRA selects the retained expert with the highest proxy fidelity as its proxy. MoRA then reuses the fitted affine map to approximate the output of pruned expert during inference.
\subsection{Preliminary}
Consider an MoE layer $\ell$ with $E_{\ell}$ routed experts.
For a token representation $h$, the router selects a top-$k_{\ell}$ expert set $\mathcal{T}_{\ell}(h)$.
The routed-expert output is:
\begin{equation}
y_{\ell}(h)
=
\sum_{e\in\mathcal{T}_{\ell}(h)}
p_{\ell,e}(h)f_{\ell,e}(h),
\label{eq:moe-output}
\end{equation}
where $f_{\ell,e}$ is the function implemented by expert $e$. $p_{\ell,e}(h)$ denotes the mixing weight applied to selected expert $e$.

\subsection{MoE pruning: router bias learning}
\label{sec:stage1}

The key challenge of MoE pruning is to retain an expert set that preserves the performance of the original MoE model. A desirable retained set should satisfy two requirements. First, it should preserve as much of the next-token prediction capability as possible. Second, the retained experts should exhibit diverse routing preferences, allowing a limited number of retained experts to cover complementary routing behaviors.

Unlike methods pruning based on statistics or reconstruction, we propose a learning-based expert pruning mechanism to efficiently identify experts that are critical to model performance. During pretraining, sparse MoE models commonly regulate expert traffic through auxiliary load-balancing objectives \citep{qwen3report2025,deepseekv22024} or expert-level correction biases \citep{liu2025muon} to mitigate routing collapse and encourage more balanced expert utilization. Building on this, we introduce an additional learnable router bias for each expert to sharpen routing probability distributions. These biases are optimized with the LM loss and a routing-diversity regularizer. The learned biases therefore can serve as retention scores that balance predictive performance and routing diversity.

\subsubsection{Learnable Router Biases}
\label{sec:priors}

MoRA augments the frozen pretrained router with one learnable router bias for each routed expert. For expert \(e\) in MoE layer \(\ell\), let \(s_{\ell,e}\in\mathbb{R}\) denote its router bias. Given the hidden state \(h\) of a token, we modify the corresponding pretrained logit as:
\begin{equation}
r'_{\ell,e}(h)
=
\frac{r_{\ell,e}(h)+\alpha_{\ell}s_{\ell,e}}{\tau},
\label{eq:adjusted-logits}
\end{equation}
where \(r_{\ell,e}(h)\) is the original router logit, \(\alpha_{\ell}>0\) controls the strength of the bias, and \(\tau>0\) is a calibration temperature. The original logit \(r_{\ell,e}(h)\) remains dependent on the current token, whereas \(s_{\ell,e}\) provides a global expert preference across all tokens. For a softmax router, let \(p_s(e\mid h)\) denote the normalized routing probability induced by the adjusted logits \(r'_{\ell,e}(h)\) before top-\(k\) selection. For any two experts \(e\) and \(e'\), their routing odds satisfy:
\begin{equation}
\frac{p_s(e\mid h)}{p_s(e'\mid h)}
=
\exp\left(
\frac{r_{\ell,e}(h)-r_{\ell,e'}(h)+\alpha_{\ell}\left(s_{\ell,e}-s_{\ell,e'}\right)}{\tau}
\right).
\label{eq:routing-odds}
\end{equation}
The learned bias therefore adjusts the router probability distributions. A larger \(s_{\ell,e}\) leads to a higher routing probability for expert \(e\) relative to other experts.

In contrast, existing statistic-based pruning methods commonly estimate expert importance using activation frequency or average routing probability \citep{koishekenov2023memory,xie2024moe}. These methods can result in a high importance score when the expert is associated with common and easily predictable tokens. However, such experts may not be critical to the model's predictive performance since these easily predictable tokens generally produce smaller language-modeling losses and weaker gradient signals during training.

\subsubsection{Optimizing the Biases for Predictive Utility and Routing Diversity}
\label{sec:prior-diversity}

MoRA learns the router biases using the LM loss on the calibration corpus. The gradient of LM loss aggregates the effect of each expert on predictive performance across calibration tokens.

Specifically, consider a softmax router with normalized selected weights, and a locally fixed top-\(k\) set \(\mathcal{T}_{\ell}(h)\). For a selected expert \(e\), the contribution through token representation \(h\) to the gradient of LM loss with respect to its bias is:
\begin{equation}
G_{\ell,e}(h)
=
\frac{\alpha_{\ell}}{\tau}
p_{\ell,e}(h)
\left\langle
g_{\ell}(h),
f_{\ell,e}(h)-y_{\ell}(h)
\right\rangle,
\label{eq:lm-bias-gradient}
\end{equation}
where \(p_{\ell,e}(h)\) is the routing weight derived from the adjusted logits \(r'_{\ell}(h)\), \(f_{\ell,e}(h)\) is the output of expert \(e\), and \(y_{\ell}(h)\) is the routed-expert mixture defined in Equation~\ref{eq:moe-output}. The vector \(g_{\ell}(h)=\partial\mathcal{L}_{\mathrm{LM}}/\partial y_{\ell}(h)\) is computed from the full batch LM objective.

A negative inner product indicates that moving this routed output toward expert \(e\)'s output locally decreases the LM loss, yielding a gradient contribution that favors increasing its bias under gradient descent. A positive inner product gives the opposite contribution. The actual bias update depends on the contributions across all tokens, the diversity regularizer, and the optimizer.

Experts with similar routing preferences may receive high biases simultaneously, limiting the diversity of routing preferences represented in the retained set under a fixed expert budget. To address this, we introduce a routing-diversity regularizer based on the geometry of the pretrained router. The regularizer suppresses the expert with the lower bias in each similar pair, encouraging the retained set to cover more distinct routing preferences. Specifically, we measure the similarity between different experts using the cosine similarity of their corresponding pretrained router rows:
\begin{equation}
C_{\ell,e,e'}
=
\left[\max\left(0,\cos\left(w_{\ell,e},w_{\ell,e'}\right)\right)\right]^2.
\label{eq:router-similarity}
\end{equation}
Here, \(w_{\ell,e}\) denotes the pretrained router row associated with expert \(e\) in layer \(\ell\). For each pair with different biases, we apply the similarity penalty only to the lower-bias expert. We define the diversity loss as:
\begin{equation}
\mathcal{L}_{\mathrm{Div}}
=
\sum_{\ell\in\mathcal{M}}
\sum_{e\neq e'}
C_{\ell,e,e'}
min(\sigma(s_{\ell,e}),\sigma(s_{\ell,e'})),
\label{eq:diversity-loss}
\end{equation}
where \(\mathcal{M}\) denotes the set of MoE layer indices. For a similar expert pair satisfying \(\sigma(s_{\ell,e})<\sigma(s_{\ell,e'})\), minimizing this loss suppresses \(\sigma(s_{\ell,e})\) without pushing down \(\sigma(s_{\ell,e'})\) through the same pair. This encourages priority separation among experts with overlapping routing preferences.

We optimize the learnable biases \(s=\{s_{\ell,e}\}\) using a joint objective that combines the LM loss with the diversity regularization:
\begin{equation}
\mathcal{L}_{\mathrm{bias}}
=
\mathcal{L}_{\mathrm{LM}}
+
\lambda_{\mathrm{d}}\mathcal{L}_{\mathrm{Div}},
\label{eq:prior-objective}
\end{equation}
where \(\lambda_{\mathrm{d}}\geq0\) controls the strength of the diversity regularization.

\subsubsection{Expert Selection from Learned Router Biases}
\label{sec:priority-pruning}

After calibration, we rank the routed experts in each MoE layer according to their learned router biases to jointly consider the predictive performance and routing diversity. Let $\mathcal{E}_{\ell}=\{1,\ldots,E_{\ell}\}$ denote the original expert set in layer $\ell$, and let $B_{\ell}$ denote the number of experts to retain. The retained and removed sets are:
\begin{equation}
S_{\ell}
=
\operatorname{Top}_{B_{\ell}}
\left(
s_{\ell,1},\ldots,s_{\ell,E_{\ell}}
\right),
\label{eq:priority-selection}
\qquad
\overline{S}_{\ell} = \mathcal{E} - S_{\ell},
\end{equation}
where $\operatorname{Top}_{B_{\ell}}$ returns the indices of the $B_{\ell}$ largest values. Experts in $\overline{S}_{\ell}$ are physically removed. The learned biases are then discarded and introduce no inference overhead.

\subsection{Post-pruning Enhancement: Expert Approximation}
\label{sec:stage2}

Although expert pruning reduces physical storage, removing experts discards their output contributions and may degrade model performance. To mitigate this degradation, MoRA fits a channel-wise affine map from each retained candidate to each removed target and evaluates its output-approximation fidelity. Each removed expert is assigned the retained candidate with the highest fidelity, and the same fitted affine map is then used to transform the selected proxy output during inference.

\subsubsection{Directed Proxy Fidelity Estimation}
\label{sec:proxy-fidelity}

We divide the calibration corpus into two non-overlapping subsets: a fit split \(\mathcal{D}_{\mathrm{fit}}\) and a scoring split \(\mathcal{D}_{\mathrm{score}}\). The fit split is used to estimate the pairwise affine maps, while the scoring split is used to evaluate their approximation fidelity. For a removed target expert \(i\in\overline{S}_{\ell}\) and a retained candidate expert \(j\in S_{\ell}\), MoRA predicts the target output using:
\begin{equation}
\widehat f_{\ell,i\leftarrow j}(h)
=
\widehat a_{\ell,i,j}\odot f_{\ell,j}(h)
+
\widehat b_{\ell,i,j},
\label{eq:affine-prediction}
\end{equation}
where \(\widehat a_{\ell,i,j}\) and \(\widehat b_{\ell,i,j}\) are the fitted channel-wise scale and bias. These parameters are estimated on \(\mathcal{D}_{\mathrm{fit}}\) using weighted ridge regression.

We derive target-conditioned token weights \(\omega_{s,t,i}\) from the pretrained router, assigning larger weights to tokens that are more relevant to target expert \(i\). Then we evaluate the affine predictor on \(\mathcal{D}_{\mathrm{score}}\) using the weighted coefficient of determination:
\begin{equation}
R^2_{\ell,i,j}
=
1-
\frac{
\sum_{(s,t)\in\mathcal{D}_{\mathrm{score}}}
\omega_{s,t,i}
\left\|
f_{\ell,i}(h_{s,t})
-
\widehat f_{\ell,i\leftarrow j}(h_{s,t})
\right\|_2^2
}{
\sum_{(s,t)\in\mathcal{D}_{\mathrm{score}}}
\omega_{s,t,i}
\left\|
f_{\ell,i}(h_{s,t})
-
\mu_{\ell,i}^{\mathrm{score}}
\right\|_2^2
},
\qquad
q_{\ell,i,j}
=
\max\left(0,R^2_{\ell,i,j}\right),
\label{eq:weighted-r2}
\end{equation}
where \(\mu_{\ell,i}^{\mathrm{score}}\) is the weighted mean output of target expert \(i\) on the scoring split. A larger \(q_{\ell,i,j}\) indicates that the affine-transformed output of retained expert \(j\) more accurately approximates the output variation of removed expert \(i\). The relation is directional because the affine map predicts \(i\) from \(j\), and both the predictor and token weights are conditioned on target expert \(i\).

Each removed expert is assigned the retained candidate with the highest proxy fidelity:
\begin{equation}
j_{\ell}(i)
=
\underset{j\in S_{\ell}}{\arg\max}\;
q_{\ell,i,j}.
\label{eq:proxy-assignment}
\end{equation}

\subsubsection{Expert Approximation based on Affine Transformation}
\label{sec:afes}

After the proxy assignment is fixed, MoRA performs the expert approximation based on affine transformation. For a removed expert \(i\) and its assigned retained proxy \(j_{\ell}(i)\), MoRA replaces the removed expert output with \(\widehat f_{\ell,i\leftarrow j_{\ell}(i)}(h)\) from Equation~\ref{eq:affine-prediction}. To limit the effect of unreliable affine approximations, we enable this proxy only when the selected proxy fidelity satisfies \(q_{\ell,i,j_{\ell}(i)}>q_{\mathrm{min}}\), where \(q_{\mathrm{min}}\) is a prescribed threshold.

No additional optimization is required after proxy assignment. The exported model stores only the retained experts, the target-proxy mappings, and the fitted channel-wise affine parameters. The removed expert parameters and proxy fidelity scores are discarded. Each pruned expert requires only \(2d_o\) affine parameters, where \(d_o\) is the expert output dimension, corresponding to one scale and one bias per output channel. This is far fewer than the number of parameters in the original expert.

\section{Experiments}
\label{sec:experiments}

\subsection{Experimental Setup}
\label{sec:experimental-setup}

\noindent\textbf{Models and pruning budgets.}
We evaluate Qwen3-30B-A3B \citep{qwen3report2025}, DeepSeek-V2-Lite \citep{deepseekv22024}, and Moonlight-16B-A3B \citep{liu2025muon}. Qwen3 has 48 MoE layers with 128 routed experts per layer and activates eight experts per token. DeepSeek-V2-Lite and Moonlight each have 26 MoE layers with 64 routed experts, two shared experts, and six activated routed experts per token. We remove either 25\% or 50\% of the routed experts in every MoE layer, retaining 96/64 experts for Qwen3 and 48/32 experts for the other two models. Shared experts and all non-expert parameters are preserved, and the original routing top-$k$ is unchanged.

\noindent\textbf{Baselines.}
\textbf{Activation Count} ranks routed experts according to how often they are selected by the pretrained router on the calibration corpus and retains the highest-ranked experts in each layer. \textbf{Average Gate Mass} ranks experts by their normalized top-\(k\) routing weights averaged over all calibration tokens, incorporating both selection frequency and routing confidence \citep{koishekenov2023memory,xie2024moe}. \textbf{MoNE} combines expert access frequency with output fluctuation to identify important experts and replaces less important experts with lightweight approximations \citep{mone2026}. \textbf{MC-SMoE} groups experts according to the similarity of their router-logit patterns and constructs representative experts through activation-frequency-weighted merging \citep{li2024merge}. \textbf{HC-SMoE} hierarchically clusters experts using their output features and merges the experts within each cluster \citep{hcsmoe2025}.

\noindent\textbf{Implementation Details.} Following the common practice in existing MoE pruning methods \citep{ling2024slimgpt,mone2026,lu2024not}, we conduct experiments using C4 \citep{raffel2020exploring} as the calibration dataset. The router biases are learned on C4 with 1000 iterations for Qwen3-30B-A3B, DeepSeek-V2-Lite and Moonlight-16B-A3B. For post-pruning enhancement, we divide the sampled C4 calibration corpus into two non-overlapping subsets of equal size: Each subset contains 64 sequences of 1024 tokens. We set the hyperparameters \(\lambda_{\mathrm{d}}=0.8\) and \(q_\mathrm{min}=0.4\).

\noindent\textbf{Evaluation benchmarks.}
Following the previous MoE pruning methods \citep{mone2026,lu2024not,ling2024slimgpt,an2024fluctuation}, we use lm-evaluation-harness \citep{gao2021framework} for zero-shot evaluation on BoolQ \citep{clark2019boolq}, RTE \citep{wang2019superglue}, HellaSwag \citep{zellers2019hellaswag}, WinoGrande \citep{sakaguchi2021winogrande}, PIQA \citep{bisk2020piqa}, OpenBookQA \citep{mihaylov2018can}, ARC-Easy \citep{clark2018think}, ARC-Challenge \citep{clark2018think}, and MMLU. HellaSwag, PIQA, OpenBookQA, ARC-Easy, and ARC-Challenge use \texttt{acc\_norm}; the other tasks use \texttt{acc}. Avg. is the mean of these nine percentages.

\subsection{Main Results}

\noindent\textbf{Comparison with baselines.}
Tables~\ref{tab:qwen-main}, ~\ref{tab:deepseek-main}, and ~\ref{tab:moonlight-main} compare MoRA with five pruning baselines on nine zero-shot benchmarks. MoRA achieves the highest average score across all three architectures at both 25\% and 50\% pruning ratios. Compared with the strongest baseline in each setting, MoRA improves the average score by 1.14, 0.43, and 1.01 percentage points at 25\% pruning, and by 6.43, 0.56, and 0.97 points at 50\% pruning, on Qwen3-30B-A3B, DeepSeek-V2-Lite, and Moonlight-16B-A3B, respectively.
We attribute the weaker performance of HC-SMoE on Qwen3 to a potential mismatch between its clustering criterion, based on average output similarity, and the high degree of expert specialization in Qwen3. Merging experts under this aggregate criterion may reduce the diversity of expert outputs.
\begin{table}[htbp]
\caption{Zero-shot results (\%) on Qwen3-30B-A3B with C4 as the calibration dataset. Bold marks the best result among MoRA and the baselines for each column at each pruning ratio.}
\label{tab:qwen-main}
\centering
\scriptsize
\resizebox{\textwidth}{!}{%
\begin{tabular}{llcccccccccc}
\toprule
Pruning ratio & Method & ARC-c & ARC-e & BoolQ & HellaSwag & MMLU & OBQA & PIQA & RTE & WinoGrande & Avg. \\
\midrule
0\% & Qwen3-30B-A3B & 56.31 & 78.87 & 88.47 & 77.66 & 77.81 & 45.00 & 80.58 & 82.67 & 70.56 & 73.10 \\
\midrule
\multirow{6}{*}{25\%}
 & Activation Count & 55.48 & 76.83 & 87.35 & 76.44 & 72.83 & 43.00 & 79.76 & 79.06 & 68.98 & 71.08 \\
 & Average Gate Mass & 55.74 & 77.12 & 87.47 & 76.56 & 72.71 & 43.60 & 80.12 & 79.12 & 69.61 & 71.34 \\
 & MoNE & 54.61 & 76.68 & 86.79 & 74.69 & \textbf{72.87} & \textbf{44.80} & 78.84 & \textbf{80.87} & 69.85 & 71.11 \\
 & MC-SMoE  & 52.52 & \textbf{80.86} & 87.38 & 77.42 & 70.62 & 43.80 & 79.82 & 79.23 & 70.59 & 71.36 \\
 & HC-SMoE  & 47.18 & 71.84 & 83.55 & 61.98 & 64.35 & 38.00 & 72.14 & 80.50 & 65.27 & 64.98 \\
 & \textbf{MoRA} & \textbf{57.42} & 80.18 & \textbf{88.81} & \textbf{79.28} & 72.32 & \textbf{44.80} & \textbf{81.18} & 76.17 & \textbf{72.38} & \textbf{72.50} \\
\midrule
\multirow{6}{*}{50\%}
 & Activation Count & 36.26 & 52.06 & 82.23 & 65.57 & 52.29 & 37.60 & 69.91 & 58.48 & 63.06 & 57.50 \\
 & Average Gate Mass & 36.26 & 54.84 & 82.66 & 66.76 & 52.98 & 36.40 & 72.03 & 58.84 & 62.04 & 58.09 \\
 & MoNE & 44.11 & 63.22 & \textbf{82.81} & 55.97 & 50.40 & 35.80 & 73.94 & 64.98 & 63.38 & 59.40 \\
 & MC-SMoE  & 23.80 & 32.37 & 75.23 & 62.49 & 22.95 & 33.20 & 60.06 & 55.60 & 62.75 & 47.61 \\
 & HC-SMoE  & 30.03 & 47.39 & 72.29 & 35.61 & 41.33 & 27.40 & 56.47 & 58.12 & 53.43 & 46.90 \\
 & \textbf{MoRA} & \textbf{46.25} & \textbf{69.07} & 82.51 & \textbf{75.20} & \textbf{53.14} & \textbf{41.80} & \textbf{80.47} & \textbf{75.09} & \textbf{68.98} & \textbf{65.83} \\
\bottomrule
\end{tabular}}
\end{table}

\begin{table}[htbp]
\caption{Zero-shot results (\%) on DeepSeek-V2-Lite with C4 as the calibration dataset. Bold marks the best result among MoRA and the baselines for each column at each pruning ratio.}
\label{tab:deepseek-main}
\centering
\scriptsize
\resizebox{\textwidth}{!}{%
\begin{tabular}{llcccccccccc}
\toprule
Pruning ratio & Method & ARC-c & ARC-e & BoolQ & HellaSwag & MMLU & OBQA & PIQA & RTE & WinoGrande & Avg. \\
\midrule
0\% & DeepSeek-V2-Lite & 49.23 & 76.60 & 79.94 & 77.92 & 54.97 & 43.80 & 80.03 & 60.65 & 71.27 & 66.05 \\
\midrule
\multirow{6}{*}{25\%}
 & Activation Count & 45.14 & 73.53 & 71.10 & 73.38 & 44.58 & 42.80 & 79.60 & 59.57 & 70.25 & 62.22 \\
 & Average Gate Mass & 45.82 & 74.41 & 70.52 & 75.26 & 46.39 & 41.80 & 79.71 & 60.29 & 69.22 & 62.60 \\
 & MoNE & 45.59 & 72.81 & \textbf{73.68} & 76.28 & \textbf{47.74} & 43.20 & 79.71 & 60.37 & 69.88 & 63.25 \\
 & MC-SMoE  & 39.08 & 66.16 & 69.20 & 67.59 & 36.20 & 37.40 & 76.99 & 57.04 & 69.30 & 57.66 \\
 & HC-SMoE  & 44.88 & 72.10 & 73.37 & 74.49 & 47.29 & 40.60 & 78.78 & \textbf{61.37} & \textbf{71.03} & 62.66 \\
 & \textbf{MoRA} &\textbf{45.99} &\textbf{77.19} &72.26& \textbf{77.24} &47.73 &\textbf{44.40}& \textbf{80.20}& 58.12& 70.01 &\textbf{63.68} \\
\midrule
\multirow{6}{*}{50\%}
 & Activation Count & 28.33 & 41.29 & 54.07 & 54.97 & 28.03 & 33.40 & 65.72 & 51.63 & 58.96 & 46.27 \\
 & Average Gate Mass & 35.24 & 56.27 & 50.58 & 67.01 & 23.46 & 36.80 & 76.06 & 49.82 & 63.62 & 50.98 \\
 & MoNE & 34.73 & 55.98 & \textbf{67.92} & 67.89 & 22.93 & 36.40 & 76.55 & 53.07 & \textbf{67.25} & 53.64 \\
 & MC-SMoE  & 28.16 & 43.98 & 57.19 & 48.56 & 22.96 & 29.60 & 66.00 & 55.23 & 60.77 & 45.83 \\
 & HC-SMoE  & 33.69 & 60.57 & 66.51 & 60.75 & \textbf{34.32} & 34.56 & 71.58 & \textbf{57.04} & 65.56 & 53.84 \\
 & \textbf{MoRA} & \textbf{38.65} & \textbf{63.17} & 58.96 & \textbf{71.47} & 23.25 & \textbf{42.20} & \textbf{77.80} & 51.63 & 62.51 & \textbf{54.40} \\
\bottomrule
\end{tabular}}
\end{table}

\begin{table}[htbp]
\caption{Zero-shot results (\%) on Moonlight-16B-A3B with C4 as the calibration dataset. Bold marks the best result among MoRA and the baselines for each column at each pruning ratio.}
\label{tab:moonlight-main}
\centering
\scriptsize
\resizebox{\textwidth}{!}{%
\begin{tabular}{llcccccccccc}
\toprule
Pruning ratio & Method & ARC-c & ARC-e & BoolQ & HellaSwag & MMLU & OBQA & PIQA & RTE & WinoGrande & Avg. \\
\midrule
0\% & Moonlight-16B-A3B & 58.19 & 82.58 & 80.15 & 78.37 & 67.31 & 45.60 & 80.96 & 66.06 & 71.43 & 70.07 \\
\midrule
\multirow{6}{*}{25\%}
 & Activation Count & 55.12 & 80.01 & 78.17 & 76.96 & 41.45 & 45.40 & 81.18 & 60.73 & 70.48 & 65.50 \\
 & Average Gate Mass & \textbf{55.24} & \textbf{80.51} & 78.32 & \textbf{77.11} & 48.29 & \textbf{46.80} & \textbf{81.50} & 60.01 & 70.43 & 66.47 \\
 & MoNE & 53.63 & 78.67 & 77.56 & 77.00 & 53.35 & 45.20 & 80.23 & 59.01 & \textbf{71.43} & 66.23 \\
 & MC-SMoE  & 48.55 & 76.73 & 78.10 & 75.27 & 39.74 & 44.80 & 80.31 & 57.40 & 70.25 & 63.46 \\
 & HC-SMoE  & 41.72 & 70.45 & 70.34 & 53.93 & 53.80 & 36.40 & 68.99 & 59.57 & 56.59 & 56.87 \\
 & \textbf{MoRA} & 53.15 & 79.12 & \textbf{79.57} & 74.76 & \textbf{56.79} & 43.00 & 80.79 & \textbf{70.76} & 69.38 & \textbf{67.48} \\
\midrule
\multirow{6}{*}{50\%}
 & Activation Count & \textbf{37.88} & 55.39 & 58.87 & 70.22 & 23.10 & 37.40 & 76.55 & 53.79 & 68.67 & 53.54 \\
 & Average Gate Mass & 37.23 & 56.26 & 62.53 & \textbf{71.46} & 23.14 & 37.40 & 76.78 & 55.40 & \textbf{68.80} & 54.33 \\
 & MoNE & 34.98 & 57.95 & 68.38 & 67.10 & 22.98 & \textbf{38.00} & \textbf{77.67} & 56.84 & 65.64 & 54.39 \\
 & MC-SMoE  & 26.71 & 43.65 & 60.49 & 46.37 & 23.18 & 29.80 & 64.69 & 52.71 & 53.35 & 44.55 \\
 & HC-SMoE  & 26.02 & 50.63 & 61.31 & 34.96 & \textbf{29.93} & 27.80 & 62.73 & 53.43 & 54.06 & 44.54 \\
 & \textbf{MoRA} & 36.43 & \textbf{65.32} & \textbf{70.92} & 61.47 & 28.66 & 35.40 & 77.31 & \textbf{61.73} & 61.01 & \textbf{55.36} \\
\bottomrule
\end{tabular}}
\end{table}

\noindent\textbf{Compression efficiency.}
As shown in Table~\ref{tab:compression-efficiency}, MoRA preserves most of the original performance at 25\% expert pruning, with average score drops of only 0.60, 2.37, and 2.59 percentage points on Qwen3-30B-A3B, DeepSeek-V2-Lite, and Moonlight-16B-A3B, respectively. The affine transformations require less than 0.08\% of the retained parameter count across all evaluated settings. Their additional inference FLOPs are bounded by \(0.0005\%\) of those of the original model, enabling post-pruning enhancement with negligible parameter and computational overhead.

\begin{table}[htbp]
\centering
\caption{Compression efficiency of MoRA. We report the model parameter counts and storage, and the additional overhead of the affine transformations.}
\label{tab:compression-efficiency}
\small
\setlength{\tabcolsep}{4pt}
\resizebox{\linewidth}{!}{%
\begin{tabular}{lccccccc}
\toprule
Model
& Pruning ratio
& Retained experts
& Model params. (B)
& Model storage (GiB)
& Affine params. (B)
& Affine storage (GiB)\\
\midrule
\multirow{3}{*}{Qwen3-30B-A3B}
& 0\%  & 128 & 30.532 & 56.871 & 0 & 0 \\
& 25\% & 96  & 23.284 & 43.371 & 0.0063 & 0.012  \\
& 50\% & 64  & 16.037 & 29.871 & 0.0126 & 0.023 \\
\midrule
\multirow{3}{*}{DeepSeek-V2-Lite}
& 0\%  & 64 & 15.706 & 29.256 & 0 & 0 \\
& 25\% & 48 & 12.108 & 22.552 & 0.0017 & 0.0032  \\
& 50\% & 32 & 8.509  & 15.849 & 0.0034 & 0.0063\\
\midrule
\multirow{3}{*}{Moonlight-16B-A3B}
& 0\%  & 64 & 15.960 & 29.728 & 0 & 0 \\
& 25\% & 48 & 12.361 & 23.025 & 0.0017 & 0.0032  \\
& 50\% & 32 & 8.763  & 16.322 & 0.0034 & 0.0063   \\
\bottomrule
\end{tabular}%
}
\end{table}

\FloatBarrier
\subsection{Ablation Study}

\begin{table}[htbp]
\centering
\small
\setlength{\tabcolsep}{4pt}
\setlength{\abovecaptionskip}{0pt}
\setlength{\belowcaptionskip}{4pt}
\caption{Component ablation on Qwen3-30B-A3B. Avg. denotes the mean score across nine benchmarks.}
\label{tab:qwen-ablation}
\begin{tabular}{lcccc}
\toprule
\multicolumn{3}{c}{Components}
& \multicolumn{2}{c}{Avg.} \\
\cmidrule(lr){1-3}
\cmidrule(lr){4-5}
\(\mathcal{L}_{\mathrm{LM}}\) & \(\mathcal{L}_{\mathrm{Div}}\) & Expert Approximation & 25\% & 50\% \\
\midrule
\(\checkmark\) & -- & -- & 67.33 & 58.08 \\
-- & \(\checkmark\) & -- & 67.48 & 54.57 \\
\(\checkmark\) & \(\checkmark\) & --
& 72.17 & 65.30 \\
\(\checkmark\) & \(\checkmark\) & \(\checkmark\)
& \textbf{72.50} & \textbf{65.83} \\
\bottomrule
\end{tabular}
\end{table}

\paragraph{Effect of the key components.}
We evaluate the effect of the key components of MoRA, containing two objectives and expert approximation. As shown in Table~\ref{tab:qwen-ablation}, jointly optimizing the LM and diversity losses achieves average scores of 72.17 and 65.30 at 25\% and 50\% pruning, respectively. Removing the diversity loss reduces these scores by 4.84 and 7.22 points, while removing the LM loss causes drops of 4.69 and 10.73 points. The larger degradation without the LM loss at 50\% pruning suggests that predictive utility becomes more important under a tighter retention budget. The diversity objective discourages overlapping routing preferences but does not assess the contribution to next-token prediction of each expert. Consequently, selecting experts solely for diversity may discard predictively useful experts, with fewer retained experts available to compensate for their removal at higher pruning ratios.
\par

\noindent\textbf{Pruning with learned biases versus adjusted routing probabilities.}

Table~\ref{tab:gate-mass-comparison} shows that computing gate mass from the bias-adjusted router improves average performance over the original gate mass in all six settings. However, ranking experts by adjusted gate mass still underperforms direct bias-based selection, particularly at 50\% pruning on Qwen3 and DeepSeek-V2-Lite, where the gaps reach 6.49 and 2.64 percentage points, respectively. These results indicate that aggregating routing probabilities after bias calibration remains less effective than directly ranking experts by the learned biases, particularly at high pruning ratios. This suggests that expert utilization, even under the calibrated router, does not fully capture predictive importance.

\begin{table}[htbp]
\centering
\small
\setlength{\tabcolsep}{7pt}
\caption{Mean scores (\%) across nine zero-shot benchmarks. Average Gate Mass uses the pretrained router, whereas \(r'\) Gate Mass uses the bias-adjusted routing probabilities. We report the average score of MoRA after pruning without expert approximation.}
\label{tab:gate-mass-comparison}
\begin{tabular}{lcccc}
\toprule
\multicolumn{1}{c}{\multirow{2}{*}{Model}}
& Pruning & Average & \(r'\) & MoRA w/o \\
& ratio & Gate Mass & Gate Mass & Expert Approximation \\
\midrule
\multirow{2}{*}{Qwen3-30B-A3B}
& 25\% & 71.34 & 71.55 & \textbf{72.17} \\
& 50\% & 58.09 & 58.81 & \textbf{65.30} \\
\midrule
\multirow{2}{*}{DeepSeek-V2-Lite}
& 25\% & 62.60 & 62.79 & \textbf{63.25} \\
& 50\% & 50.98 & 51.38 & \textbf{54.02} \\
\midrule
\multirow{2}{*}{Moonlight-16B-A3B}
& 25\% & 66.47 & 67.07 & \textbf{67.19} \\
& 50\% & 54.33 & 54.58 & \textbf{54.82} \\
\bottomrule
\end{tabular}
\end{table}

\subsection{Discussion}

\subsubsection{Learned Router Biases and Expert Deletion Sensitivity}

\label{sec:discussion-deletion}

We investigate whether learned router biases reflect the model's sensitivity to expert removal. On Qwen3-30B-A3B, we use a randomly sampled 10\% subset of the C4 calibration tokens and measure the increase in next-token prediction loss after masking one expert at a time. All pretrained parameters remain fixed, and the router recomputes top-\(k\) selection over the remaining experts.

\begin{figure}[htbp]
\centering
\includegraphics[width=0.99\textwidth]{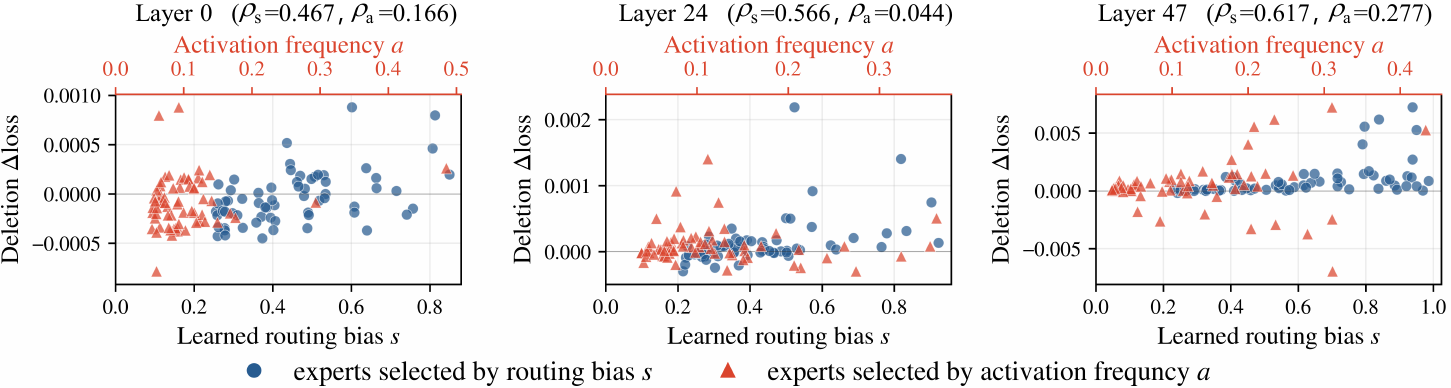}
\caption{Relationship between expert scores and deletion-induced loss increases on Qwen3-30B-A3B. Each analysis uses the top 50\% of experts within a layer ranked by the corresponding score. Spearman correlations for learned biases and activation frequency are denoted by \(\rho_s\) and \(\rho_a\), respectively.}
\label{fig:discussion-deletion}
\end{figure}

Within each layer, we analyze the top 50\% of experts ranked by learned bias and, separately, by activation frequency. Figure~\ref{fig:discussion-deletion} presents scatter plots and Spearman correlations between each score and the loss increase caused by expert deletion. Learned biases exhibit a clear positive correlation with deletion sensitivity, with stronger correlations in deeper MoE layers, whereas activation frequency shows a less consistent relationship.

\subsubsection{Token-category routing analysis}
\label{sec:discussion-token-routing}

\begin{figure}[htbp]
\centering
\includegraphics[width=0.93\textwidth]{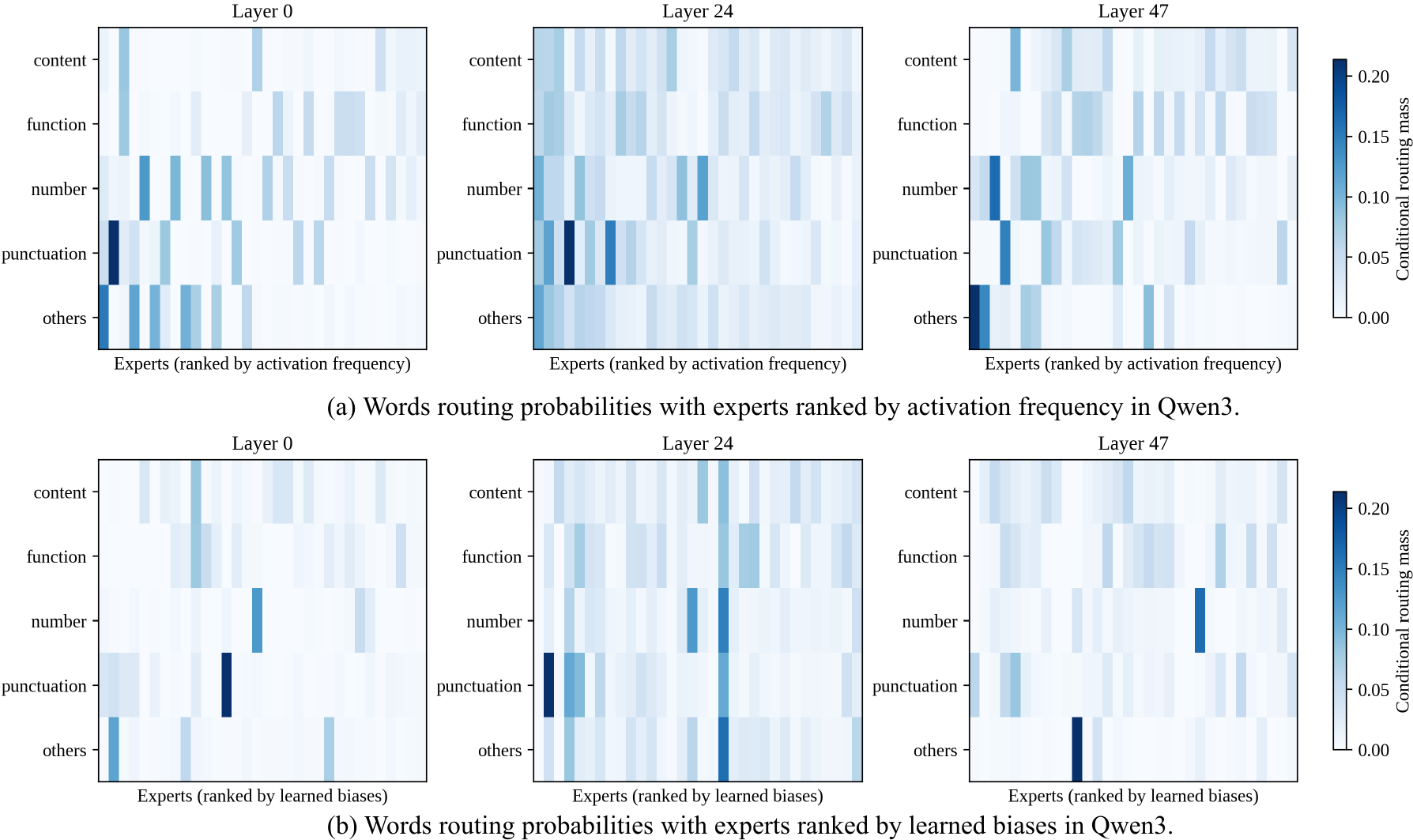}
\caption{Heatmap visualization of token-category routing probabilities in Qwen3 at Layers 0, 24,
and 47. Rows correspond to word categories, and darker cells indicate
larger routing probabilities. Experts are ordered from left to
right by decreasing activation frequency (a) and learned biases (b).}
\label{fig:figure2}
\end{figure}

We analyze token-category routing patterns in Qwen3-30B-A3B using 128 samples from C4. Specifically, we record expert assignments under the pretrained router and visualize token-category routing probabilities with experts ordered by learned bias or activation frequency. Figure~\ref{fig:figure2} shows that content words tend to be routed to experts with higher learned biases, whereas numbers and punctuation tend to be routed to experts with higher activation frequencies.

\section{Conclusion}

In this paper, we propose MoRA, a framework for structured MoE expert pruning.
We introduce a learnable router bias for each expert that sharpens the routing probability distributions. Experts that are critical to the predictive performance tend to receive higher learned biases and are therefore prioritized for retention. The retained experts are then used to approximate the outputs of the pruned experts through channel-wise affine transformations. Extensive experiments across three MoE architectures demonstrate that MoRA outperforms the evaluated baselines and preserves predictive performance with minimal additional parameter overhead.

\clearpage
\begingroup
\small
\setlength{\bibsep}{3pt plus 1pt}
\bibliography{references}
\bibliographystyle{plainnat}
\endgroup

\end{document}

%% file: math_commands.tex
\usepackage{amsmath,amsfonts,bm}

\def\eqref#1{equation~\ref{#1}}

\def\1{\bm{1}}

\DeclareMathAlphabet{\mathsfit}{\encodingdefault}{\sfdefault}{m}{sl}
\SetMathAlphabet{\mathsfit}{bold}{\encodingdefault}{\sfdefault}{bx}{n}

